\documentclass[letterpaper, 10 pt, conference]{ieeeconf}  

\IEEEoverridecommandlockouts                    

\usepackage{epsfig} 
\usepackage{amsmath} 
\usepackage{amssymb}  
\usepackage{xspace}
\usepackage{xcolor}
\usepackage{color}
\usepackage{cite}
\usepackage{booktabs, multirow}
\usepackage{colortbl}
\usepackage{bm}
\usepackage{tikz}
\usepackage{setspace}
\usepackage{graphicx}
\usepackage[ruled,vlined]{algorithm2e}
\PassOptionsToPackage{table}{xcolor}
\usepackage[font=footnotesize,labelfont=bf]{caption}
\usepackage{threeparttable} 
\newcounter{RNum}

\renewcommand{\paragraph}[1]{\vspace{0.0em}\noindent\textbf{#1}}
\newcommand{\methodname}{AnchorDream\xspace} %

\definecolor{AnchorBlue}{RGB}{15, 61, 92}
\definecolor{AnchorPurple}{RGB}{164, 109, 181}

\usepackage{gradient-text}
\newcommand{\anchordream}{
  \textsc{\gradientRGB{AnchorDream}{15, 61, 92}{164, 109, 181}}
}

\usepackage{xcolor}
\usepackage[colorlinks=true, linkcolor=black, citecolor=black, urlcolor=AnchorPurple]{hyperref}

\newcommand{\anchordreamurl}{%
  \href{https://jay-ye.github.io/AnchorDream/}{%
  https://jay-ye.github.io/AnchorDream
  }%
}
\hypersetup{
    linkcolor=AnchorPurple
}

\definecolor{myorange}{RGB}{233,113,50}
\definecolor{mygreen}{RGB}{77,167,46}

\title{\LARGE \bf
\raisebox{-0.3em}{\includegraphics[height=1.4em]{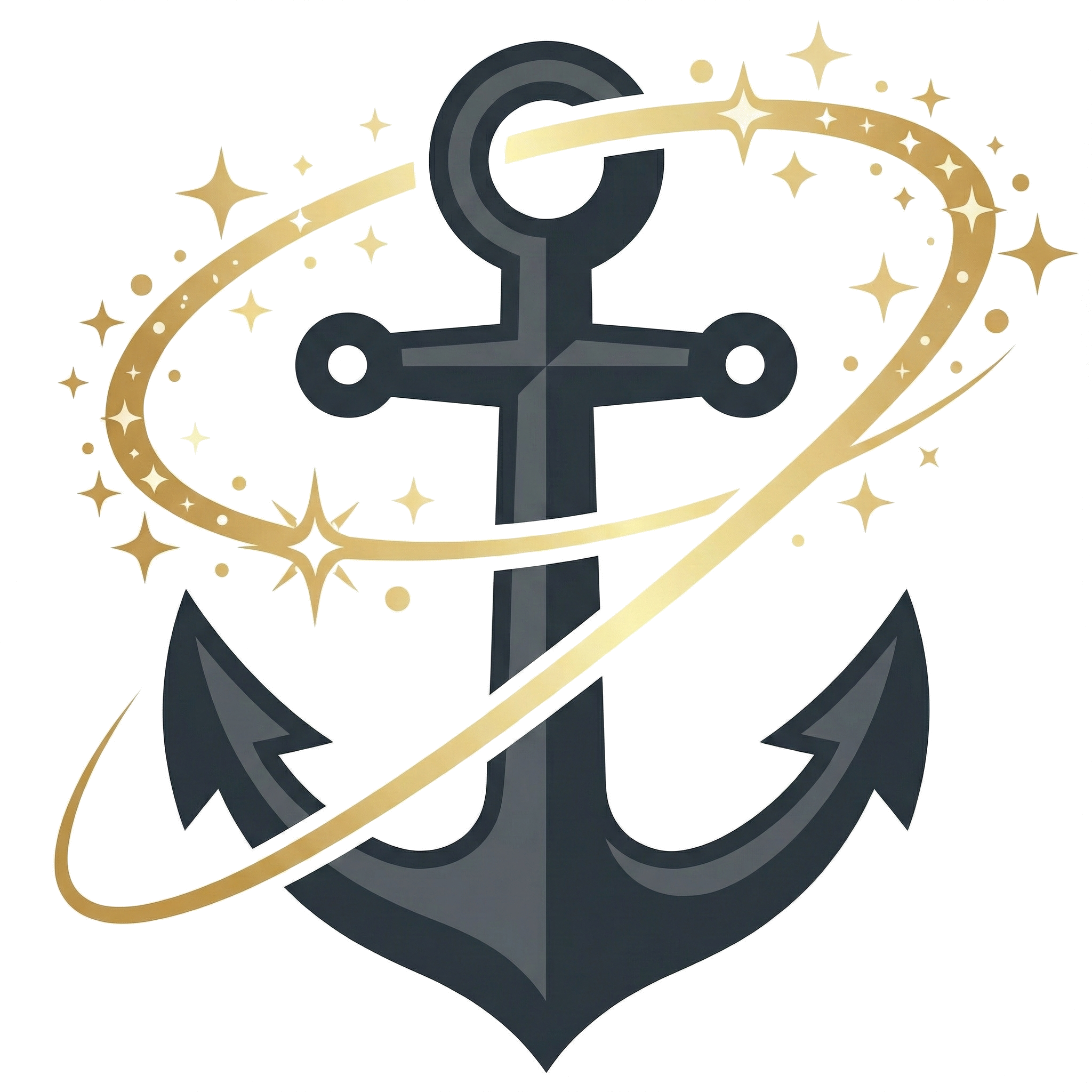}}\hspace{-2pt}%
\anchordream\\ Repurposing Video Diffusion for \\ Embodiment-Aware Robot Data Synthesis
}

\author{
    Junjie Ye\textsuperscript{$1,2$} \quad
    Rong Xue\textsuperscript{$2$} \quad
    Basile Van Hoorick\textsuperscript{$1$} \quad
    Pavel Tokmakov\textsuperscript{$1$} \quad \\
    Muhammad Zubair Irshad\textsuperscript{$1$} \quad 
    Yue Wang\textsuperscript{$2$} \quad 
    Vitor Guizilini\textsuperscript{$1$} \quad \\
    \textsuperscript{$1$}Toyota Research Institute \quad
    \textsuperscript{$2$}USC Physical Superintelligence (PSI) Lab\\
    \anchordreamurl
}

\begin{document}

\maketitle
\thispagestyle{empty}
\pagestyle{empty}

\begin{abstract}
The collection of large-scale and diverse robot demonstrations remains a major bottleneck for imitation learning, as real-world data acquisition is costly and simulators offer limited diversity and fidelity with pronounced sim-to-real gaps. While generative models present an attractive solution, existing methods often alter only visual appearances without creating new behaviors, or suffer from embodiment inconsistencies that yield implausible motions. To address these limitations, we introduce \methodname, an embodiment-aware world model that repurposes pretrained video diffusion models for robot data synthesis. \methodname conditions the diffusion process on robot motion renderings, anchoring the embodiment to prevent hallucination while synthesizing objects and environments consistent with the robot's kinematics. Starting from only a handful of human teleoperation demonstrations, our method scales them into large, diverse, high-quality datasets without requiring explicit environment modeling. Experiments show that the generated data leads to consistent improvements in downstream policy learning, with relative gains of 36.4\% in simulator benchmarks and nearly double performance in real-world studies. These results suggest that grounding generative world models\footnote{We use the term world model in a broader sense than its conventional usage in robotics and RL, where it typically refers to action-conditioned video prediction. Here, world model denotes a model that constructs coherent environments anchored to robot motion.} in robot motion provides a practical path toward scaling imitation learning.

\end{abstract}

\section{Introduction}

Imitation learning is a core approach for robotic manipulation~\cite{act, diffusionpolicy}. By training on large-scale robot demonstrations~\cite{openxembodiment2023, khazatsky2024droid}, robots can acquire complex behaviors without hand-designed rewards or task-specific controllers~\cite{lbmtri2025, Black2024_pi0, gr00tn1_2025}. However, the effectiveness of imitation learning depends critically on the scale of available data~\cite{lin2024data}. Collecting large quantities of high-quality robot demonstrations in the real world is expensive. This data bottleneck remains a major obstacle for scaling robot learning.

A growing line of work attempts to scale imitation learning by augmenting existing demonstrations from two angles. The first is to expand the \textit{observation} space~\cite{yu2023rosie, yuan2025roboengine, fang2025rebot}, altering the visual appearance of demonstrations while leaving the motions unchanged. These methods typically rely on generative models~\cite{wang2023cvpr_image_editor, Ho_VDM} to diversify scenes and objects. However, the trajectory distribution remains fixed, and no new behaviors are created. The second direction is to expand the \textit{motion} space, generating new trajectories beyond those originally collected. Such approaches, however, often depend on simulators~\cite{mimicgen}, which require labor-intensive setup and suffer from a large real-to-sim gap, or explicit scene modeling~\cite{xue2025demogen}, which limits their scalability across diverse environments.

To address these challenges, we take a different route by leveraging generative models. Video generative models~\cite{wan2025,nvidia2025cosmosworldfoundationmodel} trained on Internet-scale data, capture broad world priors including object appearances, scene layouts, and temporal consistency in motion. Unlike simulators, they require no handcrafted assets. For robotics, this suggests the potential to synthesize realistic and diverse training data at scale.
However, the challenge is \textit{embodiment grounding}.
Off-the-shelf generative models are not constrained by embodiment and often hallucinate robot bodies or produce physically inconsistent motions.
This highlights the need for a mechanism to ground these priors in real robot behavior.

\begin{figure}[!t]
\centering
  \includegraphics[width=0.95\linewidth]{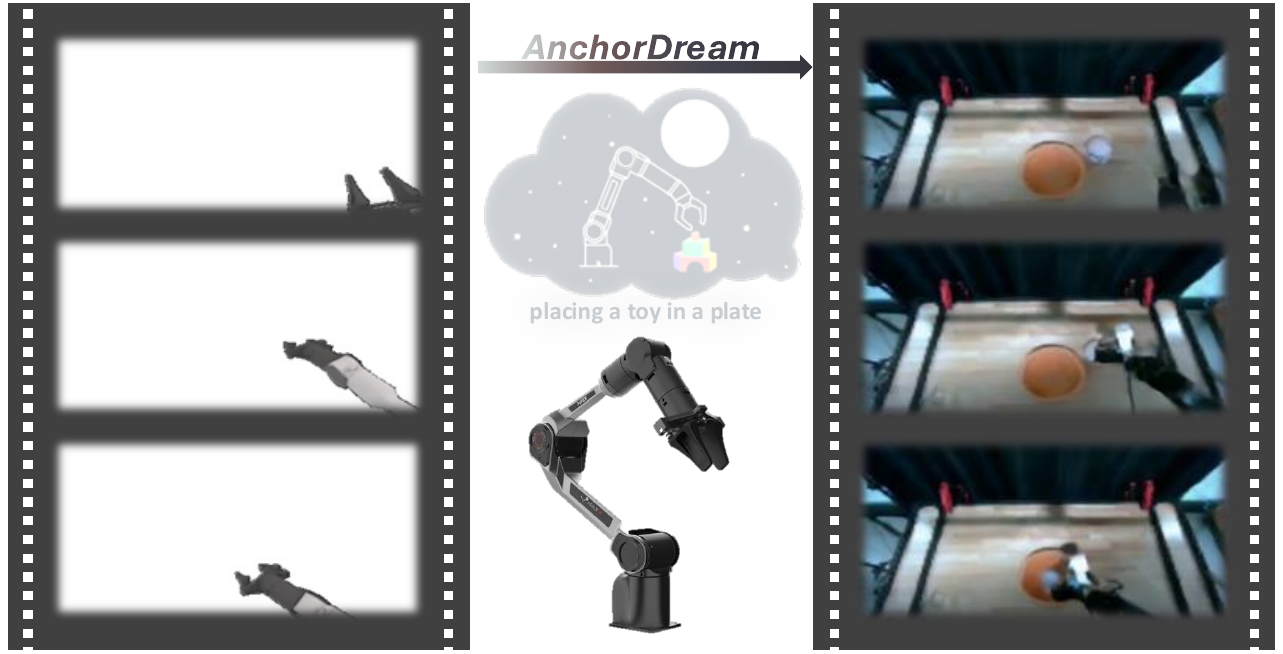}
\caption{\textbf{Overview of \methodname.} \methodname repurposes a pretrained video diffusion model as an embodiment-aware world model. Conditioned on robot motion videos, the model anchors the robot embodiment to prevent hallucination while synthesizing objects and environments consistent with the motion, enabling large-scale, high-quality demonstration generation from only a few real demonstrations.
}
\label{fig:teaser}
\vspace{-13pt}
\end{figure}

We introduce \methodname, a framework that conditions video generative models on rendered robot trajectories to synthesize demonstrations directly in the visual domain. Our approach begins with a small set of real demonstrations and then heuristically expands trajectories using perturbations of key states and motion segments to generate new trajectories at scale. Instead of reconstructing full environments in a simulator, we render \textit{only} the robot arm motions, without any scene objects or backgrounds. These trajectory replays serve as the conditioning signal for a video generative model, which synthesizes objects, interactions, and environments consistent with the observed motions. Our key idea is to decouple trajectory and environment rendering, turning actions from an afterthought into a first-class citizen. By taking control of robot trajectories and rendering them deterministically first, plausible environments and objects are generated afterwards to convey the anchoring trajectory.
This preserves embodiment consistency while producing photorealistic demonstrations that are immediately suitable for real-world policy training.

Extensive experiments in simulation and on real robots show that \methodname can expand small demonstration sets by more than an order of magnitude. The generated data leads to consistent gains in downstream policy learning, with relative improvements of 36.4\% in simulator benchmarks, and nearly doubling performance in the real-world.
These findings indicate that grounding diffusion priors in robot motion provides a practical path toward scaling imitation learning without the need for massive data collection or explicit environment modeling.

To summarize, our contributions are three-fold:
\begin{itemize}
\item We introduce \methodname, an embodiment-aware video generation framework that anchors pretrained video diffusion models in robot motion to synthesize trajectory-consistent demonstrations.
\item We propose a decoupled trajectory–environment synthesizing paradigm, where robot trajectories are expanded and rendered deterministically, and environments are generated afterwards, avoiding explicit scene modeling while preserving embodiment consistency.
\item We validate \methodname through extensive simulation and real-robot studies, showing its effectiveness for scaling imitation learning from only a handful of human demonstrations.
\end{itemize}

\begin{figure*}[!t]
\centering
  \includegraphics[width=0.95\linewidth]{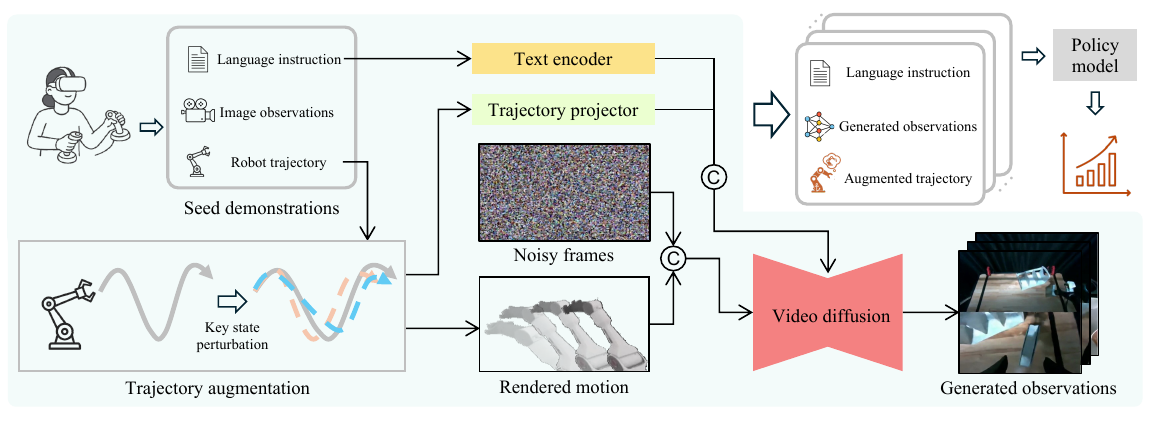}
  \vspace{-2mm}
\caption{\textbf{Outline of our proposed \methodname.} Starting from a small set of human teleoperated demonstrations, new trajectories are created by perturbing key states and recombining motion segments to ensure kinematic feasibility. Each augmented trajectory is rendered as a robot-only motion video, which, together with the task description, conditions \methodname to synthesize realistic demonstrations where environment objects are consistent with the planned trajectory. This design anchors generation on robot motion, avoiding explicit scene reconstruction and reducing the need for labor-intensive environment modeling. The synthesized demonstrations are then used to train downstream imitation learning policies, enabling limited human data to be expanded into large-scale, high-quality datasets that empower stronger policy learning.
}
\label{fig:main}
\vspace{-5mm}
\end{figure*}

\section{Related Work}
\label{sec:related_works}

Our work sits at the intersection of robot imitation learning, data augmentation, and generative models. We therefore discuss prior efforts in scaling up robotic datasets.

\subsection{Data Generation for Robot Learning}

The high cost of collecting real-world robot demonstrations has motivated two primary alternatives: simulation and data augmentation. Simulators \cite{savva2019habitat, gan2020threedworld} offer a cost-effective way to scale robot data. However, they are often limited by the diversity of available assets and suffer from the notorious sim-to-real gap. Domain randomization helps~\cite{tobin2017domain}, yet requires careful tuning and rarely covers real-world visual and physical variation. Data augmentation provides another pathway by reusing existing real demonstrations instead of collecting new ones. Early methods involved simple transformations like random cropping and color jittering~\cite{rad, yarats2021image}. More recently, generative models~\cite{nvidia_cosmos_predict2, AbuAlhaija2025CosmosTransfer1, wan2025} have enabled more advanced augmentations. A prominent line of work focuses on augmenting the \textit{observation} space while keeping the robot's actions fixed. For instance, ROSIE \cite{yu2023rosie} performs text-guided inpainting to alter objects, backgrounds, and distractors. RoboEngine~\cite{yuan2025roboengine} provides a plug-and-play pipeline combining robot segmentation with task-aware background generation. While these approaches increase visual diversity, they do not expand the underlying distribution of robot trajectories or behaviors. \methodname goes beyond visual augmentation by generating new scenes conditioned on novel robot motions, thereby diversifying both observations and behaviors.

\subsection{Synthesizing Novel Robot Trajectories}

To diversify behaviors, several approaches generate new trajectories, expanding the \textit{action} space. MimicGen \cite{mimicgen} generates new motions by composing sub-trajectories from human demonstrations and then uses a planner to execute them in a known, pre-built simulation environment to render new visual observations. This reliance on an explicit simulator, however, reintroduces the challenges of environment modeling and the sim-to-real gap. Real2Render2Real~\cite{real2render2real} follows a related idea by replaying perturbed real trajectories in simulation to generate novel demonstrations, but similarly depends on accurate environment reconstruction. DemoGen~\cite{xue2025demogen} sidesteps photorealistic rendering by operating in point clouds, recombining object-centric sub-trajectories to create obstacle-aware motions. Both expand the action distribution, but either require explicit environment modeling (e.g., simulation assets or reconstruction) or leave the pixel domain where most policies are trained.
\methodname differs in that it sidesteps the need for explicit environment assets or simulator execution altogether. By decoupling trajectory expansion from scene generation, it leverages robot motion as the sole input for synthesizing diverse and coherent demonstrations, enabling scalability without environment reconstruction.

\subsection{Generative Models for Embodied Synthesis}

Recently, there has been growing interest in using large-scale generative models, particularly video diffusion models, as implicit world models for robotics \cite{jang2025dreamgen, yuan2025roboengine}. These models, trained on vast internet datasets, possess rich priors about object physics, appearance, and temporal dynamics. DreamGen \cite{jang2025dreamgen} leverages a video model to generate entire scenes, including the robot, from a text prompt and an initial image. It then uses an inverse dynamics model to extract actions from the generated video. However, this approach faces two key challenges: 1) video models often hallucinate the robot's morphology, leading to kinematically infeasible motions, and 2) the accuracy of the extracted actions is bottlenecked by the quality of the generated video and the performance of the inverse dynamics model.

\methodname addresses these limitations with a fundamentally different conditioning scheme. Instead of generating the robot and scene jointly, we \textit{anchor} the generation process on a video of the robot's motion. By providing the robot's embodiment as a strong prior, our model is constrained to synthesize only the surrounding environment and objects in a manner consistent with the robot's kinematics. This decoupling avoids robot hallucination and bypasses the need for an inverse dynamics model, allowing us to synthesize high-quality, kinematically-grounded demonstration videos directly suitable for imitation learning.

\section{Methodology}

\subsection{Preliminaries}

\subsubsection{Video Generative Models}

Diffusion-based video generative models learn a distribution over sequences of frames by iterative denoising. 
Given a video sequence of $T$ frames $\mathbf{o}_{1:T} = \{o_1, \ldots, o_T\}$, the training objective is to approximate the conditional distribution
\begin{equation}
    p_\theta(\mathbf{o}_{1:T} \mid \mathbf{c}),
\end{equation}
where $\mathbf{c}$ denotes conditioning variables such as text, actions, or other signals. 
The model is trained to denoise a corrupted version of the sequence through a Markov chain, converging to realistic samples at inference time.  

Through large-scale training, these models encode priors on visual appearance, spatial layouts, and temporal consistency. 
For robotics, such priors can be reused to synthesize diverse and photorealistic demonstrations. 
However, they are not inherently constrained by robot embodiment, and naive generation often leads to hallucinated robot bodies, inconsistent motions, and a lack of ground truth action labels~\cite{irasim, nvidia_cosmos_predict2}. 
A promising direction is to condition the generative process on robot trajectories, which provides embodiment grounding and helps ensure that the synthesized demonstrations remain consistent with the robot’s kinematics.

\subsubsection{Procedural Trajectory Synthesis}

Let a robot trajectory be denoted as a sequence of states and actions
\begin{equation}
    \tau = \{(s_1, a_1), (s_2, a_2), \ldots, (s_T, a_T)\}.
\end{equation}

Procedural trajectory synthesis aims to expand a dataset $\mathcal{D} = \{\tau_i\}$ into a larger set $\mathcal{D}'$ by applying transformations to existing trajectories. 
MimicGen~\cite{mimicgen} is a representative approach: it segments demonstrations into object-centric subtasks, transforms them to match new scene layouts, and executes them in simulation. 
Formally, given a base trajectory $\tau$, a transformation operator $\mathcal{T}$ produces a new trajectory $\tau' = \mathcal{T}(\tau)$. 
Validation of $\tau'$ is performed in simulation to ensure task success, and successful samples are retained.  

While effective in simulation, such approaches require explicit scene modeling and physics validation, which are costly and not easily generalizable.

\subsection{\methodname}

\subsubsection{Overview}
As shown in~Fig.~\ref{fig:main}, \methodname generates large-scale demonstrations by anchoring video generative models on robot motion. 
The method separates robot trajectories from environments. 
Trajectories are expanded first, then used to condition a video model that produces photorealistic demonstrations. 
This design expands both the motion space and the observation space without requiring explicit scene modeling or access to simulation assets. 

Formally, given a small seed dataset $\mathcal{D}_0 = \{\tau_i\}$, \methodname aims to construct a dataset $\mathcal{D}' = \{(\tau', \mathbf{o}_{1:T})\}$ by synthesizing new trajectories $\tau'$ and generating corresponding observation sequences $\mathbf{o}_{1:T}$ conditioned on them.

\subsubsection{Trajectory Expansion}
We first generate new trajectories $\tau'$ by perturbing and recombining motion primitives from existing demonstrations, following prior work~\cite{mimicgen,xue2025demogen}. Specifically, key states such as contact points are manually grounded and shifted within a range. Object-centric trajectories are then stitched behind to adapt the surrounding motion, ensuring smooth transitions between segments.
This produces a large pool of trajectories that remain feasible under the robot’s embodiment. 

\subsubsection{Robot-Only Rendering}
For each synthesized trajectory $\tau'$, we render only the robot arm motion, without objects, textures, or backgrounds:
\begin{equation}
    r_{1:T} = \text{Render}(\tau'),
\end{equation}
where $\text{Render}(\cdot)$ denotes projecting the robot's 3D geometry (from its URDF/Mesh model) into a 2D image using the specified camera intrinsics and extrinsics. This produces a sequence of frames $r_{1:T}$ showing only the robot moving through the trajectory from the chosen camera viewpoints. The result is a clean, embodiment-consistent conditioning signal. 
By excluding explicit environment modeling, this step avoids the cost of replicating real-world scenes in simulation.

\subsubsection{Video Generation}
The video model takes the rendered motion traces $r_{1:T}$ and language instructions $l$, and synthesizes complete demonstrations with plausible environment and object layouts:
\begin{equation}\label{eq:videogen}
    \mathbf{o}_{1:T} \sim p_\theta(\mathbf{o}_{1:T} \mid r_{1:T}, l),
\end{equation}
where $\mathbf{o}_{1:T}$ denotes photorealistic observations consistent with $\tau'$. 
To support multi-view generation, rendered frames from different viewpoints are concatenated spatially before being passed to the model.
To make the motion trace videos compatible with pretrained large-scale video diffusion models, we concatenate them with the initialized noisy input and expand the number of input channels of the first model layer by a factor of two.
By anchoring generation on robot motion, the model preserves embodiment while filling in plausible objects and environments, thereby enabling \methodname to transform a handful of real demonstrations into large-scale, kinematically grounded datasets for imitation learning.

\begin{figure}[!b]
\centering
  \includegraphics[width=0.92\linewidth]{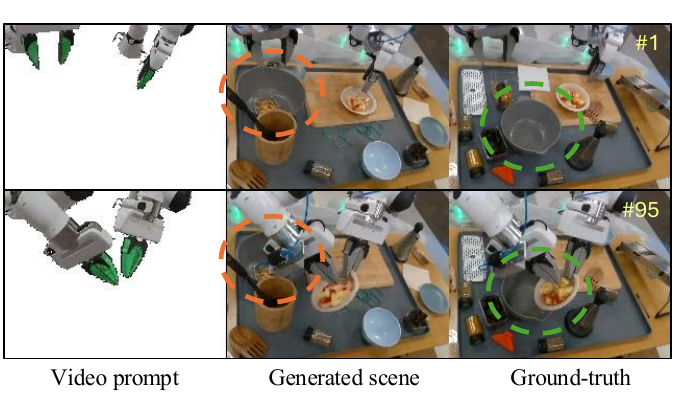}
\vspace{-6pt}
\caption{\textbf{Effect of missing global trajectory conditioning.} Without global conditioning, the generated bowl (highlighted in \textcolor{myorange}{orange}) is placed at a location that is visually plausible but not consistent with the robot’s later motion. The ground-truth bowl location (highlighted in \textcolor{mygreen}{green}) shows where the apple slices are eventually poured. This illustrates that generations based only on local context may fail to anticipate future motions.
}
\label{fig: global_traj_condition}
\end{figure}

\subsubsection{Global Trajectory Conditioning}
To generate long-horizon episodes, we autoregressively extend sequences by conditioning each new generation window on the last few frames of the previously generated clip. While this strategy allows arbitrarily long rollouts, our preliminary experiments (see Fig.~\ref{fig: global_traj_condition}) reveal that the synthesized scenes sometimes become incompatible with the robot's future motion, as the model only observes the current window and lacks awareness of upcoming waypoints. To address this issue, we provide the model with the entire trajectory $\tau'$ as an additional conditioning signal. To help the model localize the current inference window within the global context, we augment each waypoint with a binary indicator $\varphi$ marking whether it lies inside the current generation window. The global trajectory is projected into an embedding space and concatenated with the language embeddings to form the final conditioning input. The generation process in Eq.~\ref{eq:videogen} is thus reformulated as:
\begin{equation}
    \mathbf{o}_{1:T} \sim p_\theta(\mathbf{o}_{1:T} \mid r_{1:T}, l, [\tau', \varphi]),
\end{equation}
This global conditioning exposes the model to future motions and ensures that synthesized environments remain coherent over long horizons, with scene layouts aligned to the robot’s planned actions. In practice, this reduces layout drift and prevents scene–object mismatches during extended rollouts.

\begin{algorithm}[t]
\caption{Working pipeline of \methodname}
\label{alg:anchored_dreaming}
\DontPrintSemicolon
\KwIn{Seed demonstrations $\mathcal{D}_0=\{\tau_i\}$; heuristic operators $\mathcal{T}$; renderer $\mathrm{Render}(\cdot)$; video model $p_\theta$; augmentation count $K$ per seed.}
\KwOut{Augmented dataset $\mathcal{D}'=\{(\tau', \mathbf{o}_{1:T})\}$.}

$\mathcal{D}' \leftarrow \varnothing$\;

\ForEach{$\tau \in \mathcal{D}_0$}{
    \For{$k=1$ \KwTo $K$}{
        Sample operator $\mathcal{T}_k \sim \mathcal{T}$ with parameters $\phi_k$\;
        \tcp{\small{Heuristic trajectory expansion}}
        $\tau' \leftarrow \mathcal{T}_k(\tau; \phi_k)$\;

        \tcp{\small{Robot-only rendering (no scene or objects)}}
        $r_{1:T} \leftarrow \mathrm{Render}(\tau')$\;

        \tcp{\small{Trajectory-conditioned video synthesis}}
        $\mathbf{o}_{1:T} \sim p_\theta(\mathbf{o}_{1:T} \mid r_{1:T}, l, [\tau', \varphi])$\;
    }
}
\Return $\mathcal{D}'$\;
\end{algorithm}

\subsubsection{Decoupling Trajectories and Environments}
Our key idea is to decouple the two factors. 
Trajectories are fixed first, then environments are generated afterwards. 
This avoids explicit scene modeling and ensures trajectory–environment consistency. 
The final output is a dataset $\mathcal{D}' = \{(\tau', \mathbf{o}_{1:T})\}$ containing trajectory-consistent, photorealistic demonstrations. 
Each sample pairs a kinematically feasible robot trajectory with a synthesized visual sequence aligned to that motion. 
Because the data is produced directly in the visual domain, it can be used to train policies without additional transfer steps, enabling efficient scaling from a small set of seed demonstrations. A pseudocode summary of the overall data synthesis pipeline is provided in Alg.~\ref{alg:anchored_dreaming}.

\section{Experiments}
In this section, we study the following questions: (1)~Can \methodname empower better policy from a small seed set, and how close does it get to the simulator-executed upper bound? (Sec.~\ref{sec:robocasa_main}) (2)~Can policies benefit from scaling \methodname-generated data? (Sec.~\ref{sec:scale}) (3)~Which design choices of \methodname matter most? (Sec.~\ref{sec: ablation}) (4)~Does \methodname transfer to real robots and provide practical gains? (Sec.~\ref{sec:real})

\begin{table*}[!t]\centering
\caption{\textbf{Success rate comparison of policies trained with different data regimes.} \methodname consistently improves policy performance over \texttt{Human50} across all skills and approaches the policy trained with \texttt{MimicGen300}, verifying the effectiveness of anchoring video diffusion on robot motion for high-quality demonstration synthesis.}\label{tab: robocasa_main}
\resizebox{\textwidth}{!}{ 
\begin{tabular}{cccccccc|c}\toprule
&pick and place & doors & drawers & turning levers & twisting knobs &insertion & pressing buttons & Average (\%) \\ \hline
\texttt{Human50} &1.8 &31.0 &42.0 &36.0 &10.0 &12.0 &55.3 &22.5 \\
\texttt{w/ \textbf{{\methodname}300}} &4.3 &41.5 &48.0 &54.7 &21.0 &14.0 &68.7 &30.7 \\ 
\midrule
\texttt{\textcolor{gray}{w/ MimicGen300}}* &\textcolor{gray}{5.8} &\textcolor{gray}{54.0} &\textcolor{gray}{57.0} &\textcolor{gray}{64.7} &\textcolor{gray}{24.0} &\textcolor{gray}{14.0} &\textcolor{gray}{51.3} &\textcolor{gray}{33.3} \\
\bottomrule
\end{tabular}
}
\begin{tablenotes}
\footnotesize
\item *\texttt{MimicGen300} serves as an \textit{oracle} upper bound due to its reliance on privileged simulator access.
\end{tablenotes}
\end{table*}

\begin{figure*}[!t]
\centering
  \includegraphics[width=0.99\textwidth]{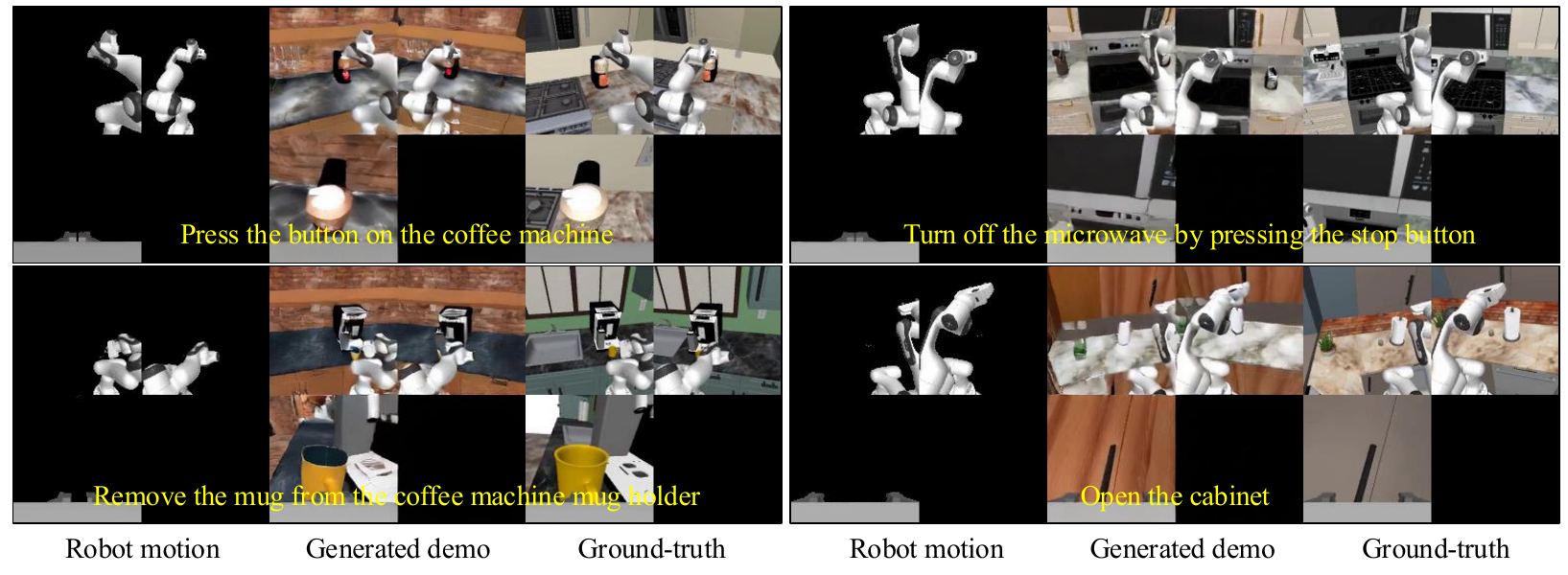}
\vspace{-2mm}
\caption{\textbf{Qualitative results on RoboCasa.} Comparison between rendered robot motion inputs, generated demonstrations, and ground-truth scenes across several tasks. The synthesized demonstrations preserve robot embodiment while producing diverse and visually coherent environments with object placements and interactions that align with the intended motions. These examples illustrate how \methodname translates abstract motion traces into realistic demonstrations, enriching the training distribution beyond the limited human demonstrations.
}
\label{fig: robocasa_vis}
\vspace{-4mm}
\end{figure*}

\subsection{Experimental Setup}
\subsubsection{Evaluation protocol}
We perform empirical evaluations of \methodname in both simulation and real-world settings. For simulation, we use the RoboCasa~\cite{robocasa2024} benchmark, which consists of 24 tabletop tasks, each with 50 human teleoperated demonstrations. RoboCasa further categorizes these 24 tasks into seven foundational manipulation skills, and we report both the average success rate within each skill and the overall average across all tasks unless otherwise noted. For real-world experiments, we design 6 everyday manipulation tasks, including placing a book on a shelf (\texttt{BookToShelf}), opening a drawer (\texttt{OpenDrawer}), closing a drawer (\texttt{CloseDrawer}), placing a toy in a plate (\texttt{ToyToPlate}), grasping and tilting a cup to pour into a bowl (\texttt{PourToBowl}), and sweeping coffee beans with a brush (\texttt{SweepCoffeeBeans}), and collect 50 human demonstrations for each using a single-arm PiPER robot. All tasks are evaluated for 50 rollouts for simulation studies and 20 rollouts for real-world evaluations.
\subsubsection{Training setup}
Unless otherwise specified, \methodname is fine-tuned from Cosmos-Predict2 2B~\cite{nvidia_cosmos_predict2} using the small set of available human teleoperation demonstrations in the simulator and real-world domains, respectively. The training is performed on 8 NVIDIA A100 GPUs over three days with LoRA~\cite{hu2022lora}. At inference, the video diffusion model generates sequences of 189 frames at a resolution of 128x128 for simulation studies and 180x320 for real-world experiments. In RoboCasa, we adopt observations from two static side-view cameras together with a wrist-mounted camera, while in real-world settings we use a third-person static camera in combination with a wrist camera. For data synthesis, we render robot-only motion videos from trajectories using RoboCasa~\cite{robocasa2024} in simulation and RoboTwin~\cite{Mu_2025_CVPR} in the real world. Importantly, \methodname leverages these simulators solely for rendering and inverse kinematics (IK) calculations, without accessing privileged environment state, simulating dynamics, or executing rollouts. Without loss of generality, we adopt BC-Transformer~\cite{robomimic} for simulator experiments and Diffusion Policy~\cite{diffusionpolicy} for real-world studies to examine the effect of \methodname demonstrations on policy learning. 

\begin{table}[!t]\centering
\small    
\caption{\textbf{Comparison of policies trained with only \texttt{Human50}, \texttt{DreamGen10K}, or \texttt{{\methodname}300}.} Training solely on \texttt{{\methodname}300} slightly surpasses \texttt{Human50} and remains competitive despite using far fewer demonstrations than \texttt{DreamGen10K}.}\label{tab: dreamgen}
\resizebox{\linewidth}{!}{ 
\begin{tabular}{ccccc}\toprule
&\texttt{Human50} & \texttt{DreamGen10K}* &\texttt{{\methodname}300} \\\hline
Average (\%) &22.5 &20.6 &\textbf{24.8} \\
\bottomrule
\end{tabular}
}
\begin{tablenotes}
\footnotesize
\item *Value taken from the original paper~\cite{jang2025dreamgen}.
\end{tablenotes}
\vspace{-4mm}
\end{table}

\subsection{How much can \methodname empower policy learning?}\label{sec:robocasa_main}
To assess whether \methodname improves policy performance from a small seed, we consider three data regimes in a multi-task training setup: 1) \texttt{Human50}: train on the 50 original demonstrations per task; 2) \texttt{w/ MimicGen300}: expands each task with 300 additional trajectories obtained by applying MimicGen’s~\cite{mimicgen} heuristic trajectory generation strategy and then executing those trajectories in the simulator to collect paired observations; 3) \texttt{w/ {\methodname}300}: use the same set of additional trajectories, but instead of executing them in the simulator, render robot-only motion videos and prompt \methodname to synthesize the corresponding observations. Since MimicGen demonstrations are realized in the simulator with privileged access to environment state, they provide the best-possible version of these trajectories, and we therefore treat \texttt{w/ MimicGen300} as an upper bound for evaluating \methodname. 

\subsubsection{Quantitative results} As shown in Table~\ref{tab: robocasa_main}, \methodname consistently improves policy performance across all skills. Training with 50 human demonstrations alone achieves an average success rate of 22.5\%, while adding 300 \methodname-generated demonstrations raises this to 30.7\%, a \textbf{36\%} relative improvement. The performance also approaches 33.3\% achieved with 300 MimicGen demonstrations, which can be regarded as an \textit{oracle} upper bound since they rely on privileged access to environment assets and simulator execution. These results verify that anchoring video diffusion on robot motion provides high-quality synthesized demonstrations that substantially empower imitation learning, narrowing much of the gap to simulator-executed data expansion without requiring explicit environment modeling.

\begin{figure}[!t]
\centering
  \includegraphics[width=0.99\linewidth]{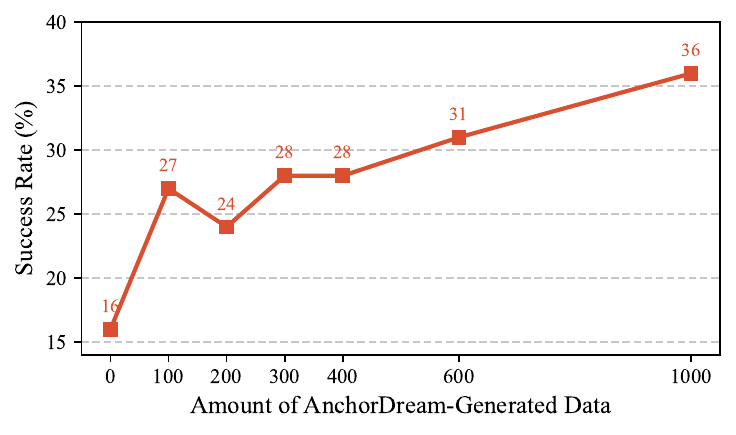}
\caption{\textbf{Effect of scaling \methodname-generated data.} Comparison of policies trained with \texttt{Human50} alone (0 on the x-axis), or with \texttt{Human50} plus different amount of \methodname-generated demonstrations on a representative subset of RoboCasa tasks. Performance improves steadily as more synthesized data are added, confirming the effectiveness of scaling \methodname for stronger policy learning.
}
\label{fig: scale}
\end{figure}

\begin{table}[!t]\centering
\caption{\textbf{Ablation on design choices in \methodname.} Comparison of policies trained with \texttt{Human50} alone, \texttt{Human50}+\texttt{{\methodname}300}, and two ablated variants without global trajectory conditioning or with a shortened inference window. Both ablations reduce performance relative to the full model, but still surpass \texttt{Human50}, verifying the robustness of \methodname for demonstration synthesis.}\label{tab: ablation}
\small
\resizebox{\linewidth}{!}{ 
\begin{tabular}{ccc}\toprule
&Average (\%) \\ \hline
\texttt{Human50} &22.5 \\\hline
\texttt{Human50} + \texttt{{\methodname}300} &\textbf{30.7} \\
\texttt{w/o global trajectory} (\ref{sec:wo_global}) &26.6 \\
\texttt{w/ shortened inference window} (\ref{sec:long_window}) &28.1 \\
\bottomrule
\end{tabular}
}
\end{table}

\begin{figure}[!b]
\centering
  \includegraphics[width=0.9\linewidth]{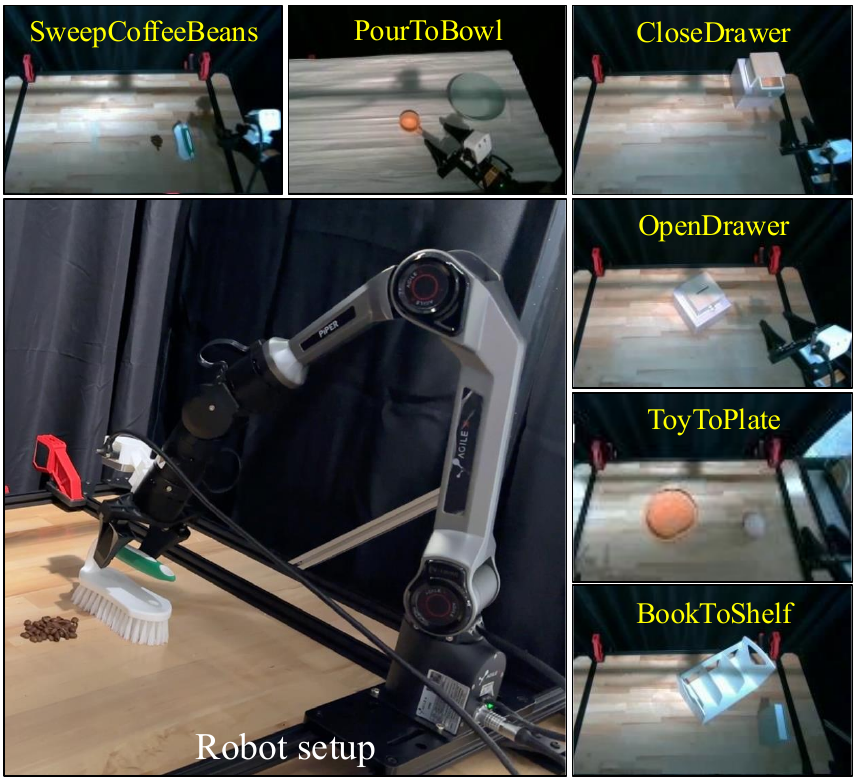}
\caption{\textbf{Real-world evaluation setup.} Six everyday manipulation tasks are used in our real-world evaluation: sweeping coffee beans with a brush, grasping and tilting a cup to pour into a bowl, closing a drawer, opening a drawer, placing a toy in a plate, and placing a book on a shelf. The lower panel shows the PiPER robot platform used for data collection and evaluation.
}
\label{fig:real_setup}
\end{figure}

\begin{table*}[!t]\centering
\caption{\textbf{Real-robot policy performance.} Comparison of policies trained with 50 human demonstrations per task (\texttt{Human50}) and with \texttt{Human50} plus $10\times$ \methodname-generated demonstrations across six everyday manipulation tasks. Augmenting with synthesized demonstrations consistently improves success rates on all tasks and raises the overall average, verifying the effectiveness of \methodname in real-world settings.}\label{tab: real}
\small
\renewcommand{\arraystretch}{1.2}
\resizebox{\textwidth}{!}{ 
\begin{tabular}{ccccccc|c}\toprule
&SweepCoffeeBeans &PourToBowl &OpenDrawer &CloseDrawer &ToyToPlate &BookToShelf &Average (\%)\\\hline
\texttt{Human50} & 35 & 0 & 0 & 30 & 85 & 20 & 28 \\
\texttt{w/ {\methodname}500} & 95 & 35 & 25 & 75 & 100 & 45 & 63 \\
\bottomrule
\end{tabular}
}
\vspace{-3mm}
\end{table*}

\begin{figure*}[!t]
\centering
  \includegraphics[width=0.99\textwidth]{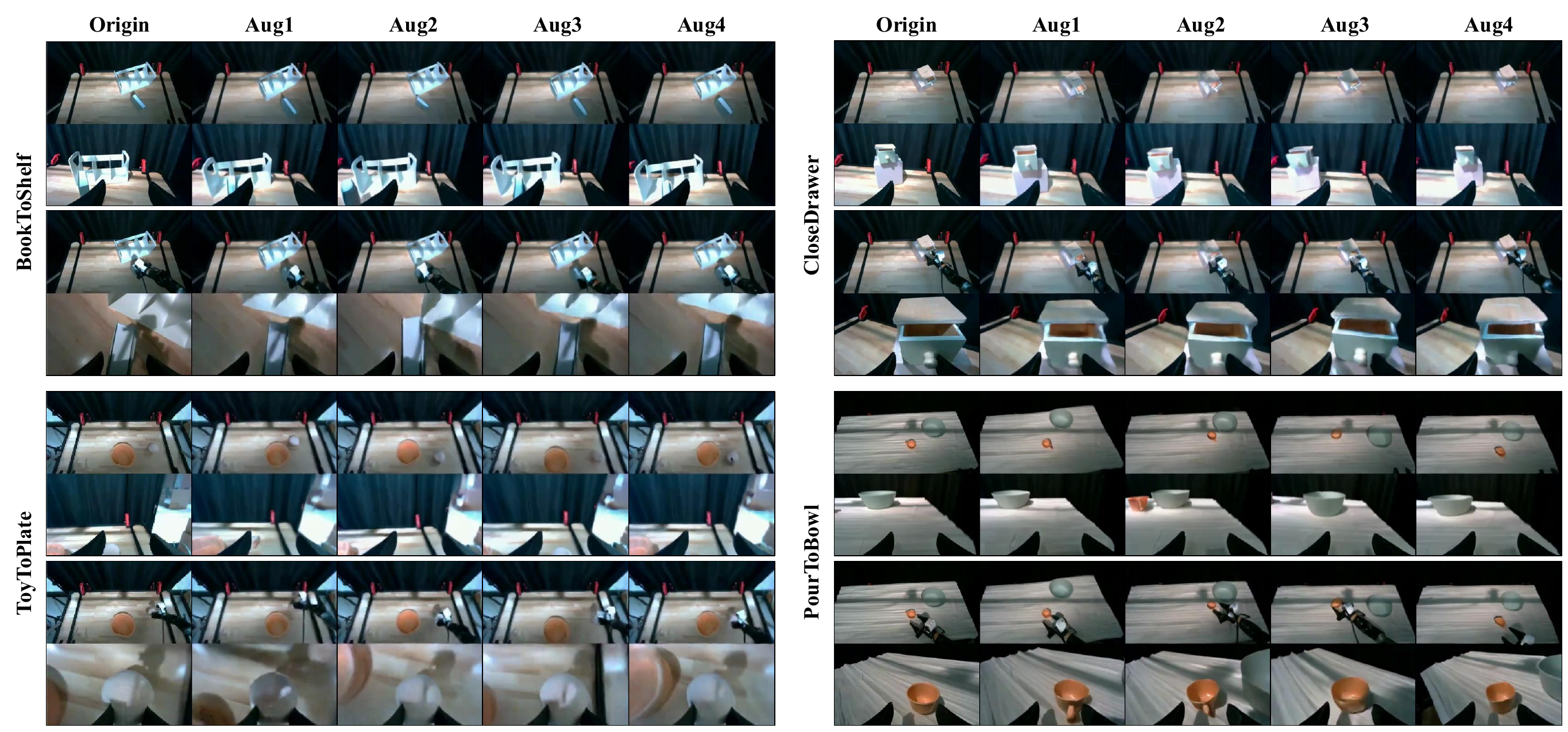}
\caption{\textbf{Real-robot qualitative results.} Example visualizations of synthesized demonstrations for several tasks. Each column shows the original trajectory (Origin) and several augmented variants (Aug1–Aug4). The generated demonstrations remain visually realistic, while the augmented trajectories steer scene layouts to diversify object positions and interactions, providing greater variability in the training data compared to the original human demonstrations.
}
\label{fig: piper_vis}
\vspace{-4mm}
\end{figure*}

\subsubsection{Qualitative results} To further illustrate the effect of \methodname, Figure~\ref{fig: robocasa_vis} presents qualitative examples comparing the input robot-only motion videos, the generated demonstrations, and the corresponding ground-truth scenes. The generated demonstrations not only preserve embodiment fidelity but also produce diverse scenes with layouts and object interactions that closely align with the ground truth. These examples confirm that the model can translate abstract motion traces into visually coherent and varied task executions, enriching the training distribution beyond what is available in the original demonstrations.

\subsubsection{Training with generated data alone} We also explore how far we can go with \methodname-generated data alone. Specifically, we train a multi-task policy using only \texttt{{\methodname}300} demonstrations and compare its performance with a policy trained on \texttt{Human50}. As shown in Table~\ref{tab: dreamgen}, policies trained solely with \methodname data slightly outperform those trained with the original 50 human demonstrations (24.8\% vs. 22.5\%). We further compare against DreamGen~\cite{jang2025dreamgen}, which generates demonstrations by producing full robot scenes with a video model and then recovering actions via an inverse dynamics model. Despite using 10k generated demonstrations per task and a foundational model~\cite{gr00tn1_2025} pretrained on large-scale robot datasets, \texttt{DreamGen10K} achieves only 20.6\% average success rate. In contrast, \methodname anchors generation on robot motion, which helps avoid embodiment hallucinations and yields demonstrations that are more consistent with downstream policy learning.

\subsection{Can scaling \methodname data help?} \label{sec:scale}
We further study whether increasing the number of synthesized demonstrations contributes to stronger policy learning. On a representative subset of seven RoboCasa tasks that cover foundational skills, we expand each task from 50 human teleoperation demonstrations to 50 plus varying amounts of \methodname-generated demonstrations, ranging from 100 up to 1000. As shown in Fig.~\ref{fig: scale}, policy performance improves steadily with more synthesized data, rising from the baseline with \texttt{Human50} to substantially higher success rates at larger scales. Despite small fluctuations at lower data sizes, the overall trend indicates that scaling \methodname data consistently boosts policy performance.

\subsection{Design Analyses}\label{sec: ablation}
We analyze two key design choices that affect long-horizon coherence and embodiment grounding in RoboCasa. 
\subsubsection{Global trajectory conditioning}\label{sec:wo_global}
The scene layout often needs to be consistent with future motion beyond the current video generation window (see Fig.~\ref{fig: global_traj_condition}). Global trajectory conditioning helps the model consider future motions while ``imagining'' the scene and object layout. As shown in the third row of Table~\ref{tab: ablation}, removing this conditioning reduces the policy success rate from 30.7\% to 26.6\%, indicating that global trajectory context is crucial for generating coherent long-horizon demonstrations.

\subsubsection{Long inference window}\label{sec:long_window}
We shorten the diffusion inference window from 189 frames to 93 frames and generate long sequences autoregressively to analyze the effect of the generation window in data synthesis. As shown in Table~\ref{tab: ablation}, the success rate decreases from 30.7\% to 28.1\%, verifying that longer inference windows are important for maintaining temporal consistency across generated sequences.

Both variants nevertheless outperform the \texttt{Human50} baseline at 22.5\%, demonstrating that \methodname remains effective even under less favorable design choices, verifying the robustness of anchoring video diffusion on robot motion for generating useful demonstrations.

\subsection{Real-Robot Evaluation} \label{sec:real}
To verify the effectiveness in real-world settings, we evaluate \methodname with six everyday manipulation tasks using a PiPER robot platform as shown in Fig.~\ref{fig:real_setup}, manually collecting 50 human demonstrations per task and fine-tuning \methodname on this data. To expand the teleoperation trajectories, we segment each trajectory into object-centric sub-trajectories following~\cite{xue2025demogen}, then randomly perturb the key states by up to $\pm10 \ \mathrm{cm}$ in the horizontal plane, render the resulting robot-only motion sequences in~\cite{Mu_2025_CVPR}, and synthesize demonstrations with \methodname, expanding each task by $10 \times$. Figure~\ref{fig: piper_vis} provides some qualitative visualizations, which show that the synthesized demonstrations are visually realistic and the augmented trajectories successfully steer generated scene layouts, enriching the training distribution.

As shown in Table~\ref{tab: real}, augmenting the human demonstrations with synthesized data leads to substantial performance gains across all six tasks. Training on the original 50 demonstrations achieves an average success rate of 28.0\%. Adding the $10\times$ \methodname-generated demonstrations raises this to 60.0\%, doubling the performance. Per-task results indicate consistent benefits. For instance, success on \texttt{SweepCoffeeBeans} improves from 35\% to 95\% and \texttt{CloseDrawer} from 30\% to 75\%. These gains confirm that the synthesized demonstrations are not only visually realistic but also effective for policy learning.

Overall, the results demonstrate that \methodname can convert a small seed set of human demonstrations into large-scale, diverse datasets that significantly empower real-robot policies. This validates the practicality of leveraging embodiment-aware video diffusion for scaling imitation learning in real-world manipulation.

\section{Conclusion}
We present \methodname, an embodiment-aware world model that repurposes pretrained video diffusion models for robot data synthesis. By anchoring generation on robot motion, \methodname produces kinematically grounded and visually realistic demonstrations, enabling scalable imitation learning without explicit environment modeling or simulator rollouts. 
These results verify the effectiveness of anchoring video diffusion on robot motion as a practical path to large-scale policy learning and point toward integrating embodiment priors with generative models to expand diversity and usability of synthesized robot data.
While our study focuses on tabletop manipulation tasks, extending \methodname to broader domains such as mobile or long-horizon manipulation offers an exciting avenue for future work.

\section{Acknowledgment}
We thank our friends and colleagues, including Jiawei Yang, Chen Xu, Masha Itkina, and Mingtong Zhang, for their helpful discussions and insightful suggestions. This work was partially supported by the National Science Foundation through NSF CPS \#2434460. The USC Physical Superintelligence Lab acknowledges generous support from Toyota Research Institute, Dolby, Google DeepMind, Capital One, Nvidia, and Qualcomm. Yue Wang is also supported by a Powell Research Award.

\bibliographystyle{IEEEtran}
\bibliography{icra25}

\end{document}